\documentclass{article}

\usepackage{PRIMEarxiv}

\usepackage[utf8]{inputenc} 
\usepackage[T1]{fontenc}    
\usepackage{url}            
\usepackage{booktabs}       
\usepackage{amsfonts,amsmath,amssymb,amsthm}
\usepackage{nicefrac}       
\usepackage{microtype}      
\usepackage{fancyhdr}       
\usepackage{graphicx}       
\usepackage{algorithm}
\usepackage{algpseudocode}
\usepackage{adjustbox}
\usepackage{xspace}
\usepackage{multirow}
\usepackage{xcolor}
\usepackage{tabularx}
\usepackage{array}
\usepackage[numbers,sort&compress]{natbib}
\usepackage{placeins}
\usepackage{hyperref}       

\graphicspath{{Figures/}{./}}

\newcommand{\promptbox}[1]{%
  \par\smallskip\noindent
  \fbox{\begin{minipage}{\dimexpr\linewidth-2\fboxsep-2\fboxrule\relax}
  \footnotesize\ttfamily\raggedright\sloppy
  #1
  \end{minipage}}\par\smallskip
}

\title{Verify--Backtrack Loop: Structured Precondition Recovery for Long-Horizon Language Agents}

\author{
  Mingyang Liu$^{1,*}$,
  Qingcan Kang$^{2,*}$,
  Shixiong Kai$^{2}$,
  Tao Zhong$^{2}$, \\
  Mingxuan Yuan$^{2}$,
  Xin Li$^{2}$,
  Linqi Song$^{1,\dagger}$ \\[0.8em]
  {\normalfont\normalsize
  \begin{tabular}{c}
    $^{1}$Department of Computer Science, City University of Hong Kong \\
    $^{2}$Huawei Noah's Ark Lab, Hong Kong \\[0.35em]
    $^{*}$Equal contribution \quad
    $^{\dagger}$Corresponding author
  \end{tabular}}
}

\begin{document}
\maketitle

\begin{abstract}
Language agents increasingly undertake long-horizon tasks in interactive environments, where actions often depend on latent preconditions. The core difficulty is not trajectory length but recovery: when a prerequisite is missing or an earlier decomposition becomes invalid, substituting another plausible action rarely helps. ReAct and subsequent reflection or adaptive-planning methods support local correction, but recovery is expressed as free-form text, leaving no explicit control state that links the terminal goal, the active goal, and the source of failure. We introduce the Verify--Backtrack Loop (VBL), a training-free extension of ReAct that converts precondition discovery and failure diagnosis into structured updates of planning state. VBL maintains the evolving task decomposition as a bounded goal stack and, whenever a readiness check is due, verifies two horizons jointly---whether the active goal contributes to the terminal goal and whether it is feasible in the current state---on goal activation, after execution failure, and after a material state change. These dual-horizon judgments, combined with typed failure attribution, trigger explicit control operations---\textsc{Push}, \textsc{Pop}, retry, or \textsc{BacktrackTo} the parent. We evaluate VBL with Qwen3.5-9B, GPT-4o, and DeepSeek-V3.2 on three public benchmarks: AppWorld, ALFWorld, and ScienceWorld. VBL improves over ReAct and reflection-based control on every backbone--benchmark pair. On the challenging AppWorld setting, it improves task success by up to 22.07 percentage points over ReAct, with gains extending to ALFWorld and ScienceWorld. Ablations show that removing either verification or diagnosis degrades performance---in several settings to below the ReAct baseline---so the two mechanisms are complementary rather than redundant. These results demonstrate that reliable long-horizon interaction benefits from revising the planning state, not merely the next action.
\end{abstract}


\section{Introduction}
\label{sec:intro}

Language models increasingly act through tools, web interfaces, software
environments, and embodied systems~\citep{yao2023react,huang2022inner}. Many of
these interactions are long-horizon: actions change an external state, their
effects accumulate, and later decisions depend on earlier outcomes. Reliable
execution therefore requires an agent to preserve the terminal goal while
adapting its intermediate plan. The difficulty is not simply trajectory length.
Dependencies among intermediate goals may be absent from the instruction and
become visible only during execution. When such a prerequisite is missed,
several subsequent actions may fail. Choosing another plausible action is
rarely enough: the agent must identify the source of the failure and revise the
corresponding level of task decomposition.

ReAct~\citep{yao2023react} provides a general basis for interactive agents by
interleaving reasoning, action, and observation. It adapts the next decision to
new evidence, but it does not impose an explicit hierarchy over goals. Its
interaction history records what happened without specifying which
short-horizon goal is active, how that goal contributes to the terminal
goal, or where planning should resume after repeated failure. Adaptive and search-based planners revise plans or explore alternative
paths~\citep{sun2023adaplanner,hao2023reasoninglanguagemodelplanning,zhou2024languageagenttreesearch},
and reflection methods revise outputs or actions through critique and
goal-conditioned feedback~\citep{madaan2023selfrefine,shinn2023reflexion,reflact2025}.
These approaches improve adaptation, yet recovery remains predominantly
textual: there is typically no compact planning state that jointly represents
the terminal goal, the active goal, and the causal source of a failure.

The central challenge is therefore not merely to revise the next action, but to
revise the planning state at the appropriate level. Such revision requires two
goal horizons. The \emph{terminal goal} defines task completion, whereas the
\emph{active goal} defines the immediate objective under the current
decomposition. Before execution, the active goal should both contribute to the
terminal goal and be feasible in the current state. After failure, the agent
must determine whether to insert a missing prerequisite, retry locally, or
return to an earlier decomposition.

\begin{figure}[ht]
    \centering
    \includegraphics[width=0.72\linewidth]{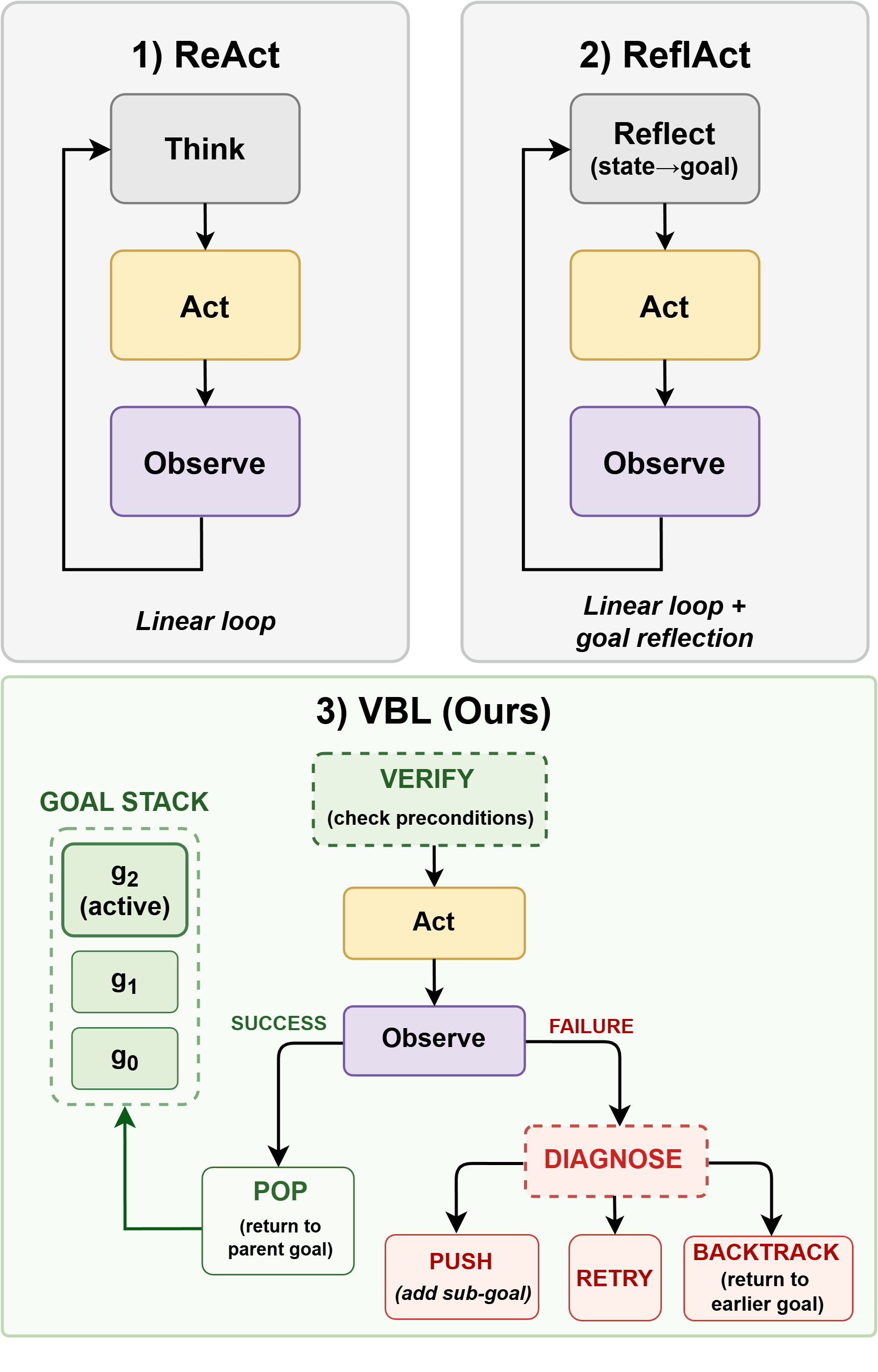}
    \caption{Control structures compared. (1) ReAct interleaves thought,
    action, and observation in a linear loop. (2) ReflAct replaces the thought
    with reflection on the current state relative to the task goal, but the
    loop stays linear. (3) VBL maintains an explicit goal stack:
    \textsc{Verify} checks alignment and readiness when $\mathit{due}$ is set
    (on activation, after failure, or after a material state change), a
    completed goal is removed with \textsc{Pop}, and a failure is routed
    through \textsc{Diagnose} to one of \textsc{Push} (insert a prerequisite),
    a local retry, or \textsc{BacktrackTo} (resume from the parent).
    The executor receives the root task text together with the active
    goal---here $g_2$---not the full stack.}
    \label{fig:framework}
\end{figure}

Based on this formulation, we propose the \textbf{Verify--Backtrack Loop
(VBL)}, a training-free extension of ReAct that converts precondition discovery
and failure diagnosis into structured updates of planning state. VBL represents
the evolving decomposition as a bounded goal stack whose root is the terminal
task and whose top is the active goal. A dual-horizon verifier evaluates
terminal-goal alignment and active-goal readiness when the active goal is
installed, after an execution failure, and after a material state change.
Typed failure attribution then maps repeated failures to explicit stack
operations---inserting a missing prerequisite with \textsc{Push}, or
\textsc{BacktrackTo} the parent---while a goal verified as complete is removed
with \textsc{Pop}. Because each stack update re-arms verification, a restored
ancestor that is still misaligned triggers further steps upward.
The executor is conditioned on the root
task text and the selected active goal, so inserting a prerequisite or returning
to a parent directly changes subsequent action selection. Backtracking changes
the planning state rather than the environment state: executed actions are not
reversed; instead, invalid descendants are removed and planning resumes from the
restored ancestor under the current observation. This distinction makes VBL
applicable to stateful environments in which actions may be irreversible.
Figure~\ref{fig:framework} contrasts this control structure with the linear
loops of ReAct and ReflAct.

We evaluate VBL with Qwen3.5-9B, GPT-4o, and DeepSeek-V3.2 on three public
benchmarks: AppWorld, ALFWorld, and ScienceWorld. Holding the backbone,
interaction budget, and environment interface fixed within each setting, VBL
improves over the baselines on every backbone--benchmark pair.
Relative to ReAct, it improves task success by up to 22.07 percentage points on
AppWorld, 14.21 percentage points on ScienceWorld, and 10.45 percentage points
on ALFWorld. Component ablations further show that removing either verification
or diagnosis degrades performance, in several settings to below the ReAct
baseline, so the two mechanisms are complementary rather than redundant: a goal
stack driven by only one of them can be worse than no stack at all.
Our contributions are as follows:
\begin{itemize}
  \item We formulate long-horizon agent control as a dual-horizon problem that
  distinguishes terminal-goal alignment from active-goal readiness, allowing
  misaligned and blocked goals to trigger different planning decisions.
  \item We propose VBL, a training-free extension of ReAct that maps
  verification and typed failure attribution to explicit transitions over
  a bounded goal stack.
  \item We evaluate VBL across three public interactive benchmarks and three
  backbone models. With backbone, budget, and environment interface held fixed,
  VBL improves task success over the baselines on every evaluated pair, while
  ablations show that verification and diagnosis are complementary rather than
  redundant.
\end{itemize}

\section{Related Work}
\label{sec:related}

\paragraph{Reasoning, acting, and reflection.}
Chain-of-thought prompting elicits intermediate reasoning for multi-step
problems~\citep{wei2022cot}, and ReAct extends this principle to interactive
settings by interleaving reasoning, action, and
observation~\citep{yao2023react}. Feedback-based variants revise outputs or
actions through critique and reflection: Self-Refine iterates on model
outputs~\citep{madaan2023selfrefine}, Reflexion stores verbal lessons across
attempts~\citep{shinn2023reflexion}, and ReflAct conditions the next action on
the relation between the current state and the task goal~\citep{reflact2025}.
These methods show that reflection improves adaptation, but recovery typically
remains free-form text rather than an update to an explicit control state. VBL
addresses that interface by separating terminal-goal alignment from active-goal
readiness and mapping verification and typed failure attribution to prescribed
transitions over a goal stack. We compare against ReAct and ReflAct under a
shared executor; multi-trial memory and multi-stage metacognitive controllers
are outside our single-episode protocol.

\paragraph{Planning, decomposition, and search.}
Prompted decomposition~\citep{zhou2023leasttomost,wang2023planandsolve} and
adaptive or search-based planners~\citep{sun2023adaplanner,yao2023tot,%
besta2024got,hao2023reasoninglanguagemodelplanning,zhou2024languageagenttreesearch}
revise an explicit plan or explore alternative reasoning and action paths. VBL
asks a narrower within-episode question: when a single active path becomes
blocked, which level of its decomposition should change? It retains one bounded
goal stack rather than a search tree, inserts a missing prerequisite when the
active goal is aligned but infeasible, and backtracks toward the parent when the
current decomposition is invalidated. The emphasis is on explicit recovery
points and targeted planning-state revision rather than broad path exploration.

\paragraph{Hierarchical control and agent memory.}
Hierarchical agents organize long trajectories around subgoals; HiAgent, for
example, treats subgoals as working-memory units and summarizes observations
tied to completed subgoals~\citep{hu2024hiagenthierarchicalworkingmemory}.
VBL's hierarchy is the bounded goal stack: it represents the current
decomposition and its recovery points, and the executor is conditioned on the
root task text and the active goal rather than the full stack.
Episode-local summaries used by the full VBL configuration
(Section~\ref{subsec:bounded-execution}) are an implementation detail for
limiting context rather than a separate memory contribution; they are discarded
when the task ends.

VBL's distinguishing contribution is therefore not any single ingredient above,
but their coupling through a common control representation: a goal-stack
transition that links terminal alignment, local feasibility, failure
attribution, and subsequent execution.

\section{Methodology}
\label{sec:method}

VBL augments a ReAct executor with an explicit planning state. Its central
principle is to separate two decisions that are often merged in a free-form
trajectory: whether the active goal is both relevant and feasible, and how
the task decomposition should change when execution fails. A bounded goal stack
turns these judgments into control operations.
Figure~\ref{fig:framework} contrasts this control structure with the linear
loops of ReAct and ReflAct.

\subsection{Problem Formulation}

We consider an agent that receives a natural-language instruction and acts in a
stateful environment. A task is represented as a goal-conditioned, partially
observed sequential decision process
\[
  \mathcal{M}=\bigl(\mathcal{S},\mathcal{A},\mathcal{O},\mathcal{T},
  \Omega,I,g^\star,B\bigr).
\]
Here, $\mathcal{S}$, $\mathcal{A}$, and $\mathcal{O}$ denote the state, action,
and observation spaces. The transition function $\mathcal{T}$ governs changes
to the environment, and $\Omega$ produces an observation after each action. The
instruction $I$ induces a terminal goal predicate
$g^\star:\mathcal{S}\rightarrow\{0,1\}$. The interaction budget is $B$.

The true state $s_t$ is not directly observed. Let $o_0$ denote the initial
observation before any action, and let $o_t$ ($t\geq 1$) denote the observation
returned after executing $a_t$. At decision step $t\geq 1$, the agent conditions
on the interaction history
\[
  h_t=(I,o_0,a_1,o_1,\ldots,a_{t-1},o_{t-1}),
\]
with $h_1=(I,o_0)$. Before selecting $a_t$, the controller reads the latest
available observation
$o^{\mathrm{curr}}_t:=o_{t-1}$ when $t>1$ and $o^{\mathrm{curr}}_1:=o_0$.
Post-action observations may carry execution feedback for the preceding
action---for instance an error message or a failed unit test---which is the
only failure signal the agent receives.
An episode $\tau$ succeeds if the terminal goal is reached within budget:
\[
  R(\tau)=\mathbb{1}\!\left[\exists t\leq B:
  g^\star(s_t)=1\right].
\]
Long-horizon tasks contain intermediate goals whose dependencies are not fully
specified by the instruction. Let
$\Psi=\{\psi_1,\ldots,\psi_m\}$ denote a latent set of subgoal predicates, with
$g^\star\equiv\psi_m$, and let $\prec$ denote a latent precedence relation. If
$\psi_i\prec\psi_j$, then $\psi_i$ must hold before an action can complete
$\psi_j$:
\[
  \Pr[\psi_j(s_{t+1})=1\mid s_t,a_t]>0
  \Longrightarrow
  \psi_i(s_t)=1.
\]
Neither $\Psi$ nor $\prec$ is given to the agent. Both must be inferred from the
instruction, the interaction history, and environmental feedback. The predicate
$g^\star$ is likewise unavailable to the agent; it is evaluated only by the
environment to score an episode. Every judgment VBL makes internally is
conditioned on the instruction text $I$, which is the agent's only
representation of the terminal goal, together with $h_t$ and
$o^{\mathrm{curr}}_t$.

\subsection{Dual-Horizon Control State}
\label{subsec:control-state}

VBL represents the task at two temporal scales. The terminal goal $g^\star$
defines successful completion. The active goal $g_t$ defines the immediate
objective under the current decomposition. These goals serve different roles.
Terminal-goal alignment prevents local progress from drifting away from the
task, whereas active-goal readiness prevents the agent from attempting a step
whose prerequisites are not yet satisfied.

At each decision point, VBL estimates two properties:
\[
  \operatorname{Align}(g_t,g^\star\mid h_t),
  \qquad
  \operatorname{Ready}(g_t\mid h_t,o^{\mathrm{curr}}_t).
\]
Alignment concerns the contribution of $g_t$ to $g^\star$. Readiness concerns
the feasibility of $g_t$ in the current environment. Both are properties of the
underlying task, so VBL only estimates them; as noted above, the estimates are
formed from the instruction text rather than from $g^\star$ itself. Keeping these judgments
separate allows a goal to be represented as aligned but blocked, or feasible
but misaligned. The former calls for a prerequisite; the latter calls for
revision of the decomposition.

The planning state is a bounded goal stack $S_t$. Its root $g_0$ stores the
instruction text that induces the terminal predicate $g^\star$, and its top
element is the active goal $g_t$.
Each node stores a description $d(g)$, a parent pointer, a consecutive-failure
counter $n_{\mathrm{err}}(g)$, a retry counter $n(g)$, and a set $P(g)$ of
previously proposed prerequisites; a path may additionally be marked failed
when recovery abandons it.

The stack supports three operations:
\begin{itemize}
  \item $\textsc{Push}(p)$ places a missing prerequisite above the active goal;
  \item $\textsc{Pop}()$ marks the active goal as complete and restores its
  parent as the active goal; and
  \item $\textsc{BacktrackTo}(g')$ removes the descendants of an ancestor $g'$
  and resumes decomposition from $g'$. In the controller, each invocation
  targets the parent of the active goal; because every stack update re-arms
  verification, a parent that is itself misaligned triggers a further step
  upward, so ascent continues until an aligned ancestor remains or the root
  $g_0$ is reached. All later mentions of backtracking assume this behavior.
\end{itemize}
Alignment holds trivially at the root, since $d(g_0)=I$ is the terminal task
description itself; \textsc{Verify} there only separates a completed task from an
incomplete one, so ascent always terminates at the root. Backtracking at the
root, which diagnosis can still prescribe, invalidates the current decomposition
but retains $g_0$. The operation changes planning state only. It does not reverse
actions or restore an earlier environment state. Replanning therefore conditions on the
current observation, including all effects of earlier actions.

Each control iteration optionally verifies the active goal and then executes it
with the ReAct policy. If the goal is completed, VBL pops it; if execution repeatedly
fails, VBL diagnoses the cause and applies one stack transition before resuming.
Algorithm~\ref{alg:vbl} states this control skeleton; admissibility checks,
retry/backtrack budgets, episode-buffer updates, and the full branching logic
used in experiments appear in Appendix~\ref{app:alg}.
The ReAct executor retains its original instruction and action format. At step
$t$, VBL supplies the decision context
\[
  c_t=\bigl(d(g_0),d(g_t),\operatorname{Retrieve}(E_t,g_t)\bigr),
  \qquad
  a_t\sim\pi_\theta(\cdot\mid h_t,c_t),
\]
where $d(g_0)=I$ is the root instruction that induces $g^\star$, and $E_t$ is a
lightweight episode buffer (Section~\ref{subsec:bounded-execution}).
Thus, the
executor observes the terminal-task text, the selected active goal, and only the
records relevant to that goal; the remaining stack stays inside the controller.

\begin{algorithm}
\caption{Verify--Backtrack Loop (simplified)}
\label{alg:vbl}
\begin{algorithmic}[1]
\Require environment $\mathcal{M}$, ReAct policy $\pi_\theta$, budget $B$,
threshold $K$
\State $S\gets[g_0]$ with $d(g_0)\gets I$; $t\gets0$
\While{$S\neq\varnothing$ \textbf{and} $t<B$}
  \State $g\gets\mathrm{top}(S)$
  \If{$g$ needs verification}
    \Comment{on activation; after failure; after material state change}
    \State $(q,u,p,e)\gets\textsc{Verify}(d(g_0),g,\ldots)$
    \If{$q=\mathrm{complete}$} $\textsc{Pop}()$; \textbf{continue}
    \ElsIf{$q=\mathrm{misaligned}$}
      $\textsc{BacktrackTo}$ parent; \textbf{continue}
      \Comment{re-verify the restored ancestor next}
    \ElsIf{$u=\mathrm{blocked}$ \textbf{and} $p$ admissible}
      $\textsc{Push}(p)$; \textbf{continue}
    \EndIf
  \EndIf
  \State $a\sim\pi_\theta(\cdot\mid h,c)$; execute $a$; observe $o'$; $t\gets t+1$
  \If{$g$ complete under $o'$} $\textsc{Pop}()$; \textbf{continue} \EndIf
  \If{$o'$ is a failure}
    \State $n_{\mathrm{err}}(g)\gets n_{\mathrm{err}}(g)+1$
    \If{$n_{\mathrm{err}}(g)\geq K$}
      \State $(z,p,e)\gets\textsc{Diagnose}(d(g_0),g,o',\ldots)$;
      $n_{\mathrm{err}}(g)\gets0$
      \State apply one of $\textsc{Push}$ / retry /
      $\textsc{BacktrackTo}$(parent) according to $z$
    \EndIf
  \EndIf
\EndWhile
\end{algorithmic}
\end{algorithm}

\subsection{Verification and Structured Recovery}
\label{subsec:recovery}

Verification is gated by a flag $\mathit{due}$, raised on goal activation
(including after $\textsc{Push}$, $\textsc{Pop}$, or $\textsc{BacktrackTo}$),
after failure, and after a material state change, and cleared once
\textsc{Verify} runs. Between actions that leave readiness intact, the
controller does not re-verify. Each call uses the same underlying model under a
fixed output schema; its tokens are counted like executor calls. Prompts use
$d(g_0)=I$ in place of $g^\star$. The verifier returns
\[
  \textsc{Verify}(d(g_0),g_t,o^{\mathrm{curr}}_t,h_t,E_t)
  \rightarrow (q_t,u_t,p_t,e_t),
\]
with alignment
$q_t\in\{\mathrm{aligned},\mathrm{misaligned},\mathrm{complete}\}$,
readiness
$u_t\in\{\mathrm{ready},\mathrm{blocked},\mathrm{unknown}\}$, an optional
missing prerequisite $p_t$, and evidence $e_t$.

Three rules follow. A completed goal is popped. A misaligned goal triggers
$\textsc{BacktrackTo}$ its parent and re-arms $\mathit{due}$, so the restored
ancestor is re-verified before the next action. An aligned but blocked goal
pushes $p_t$ when admissible---concrete, actionable, absent from both $P(g_t)$
and the stack, and within depth and push limits. If readiness is uncertain, or
if the goal is blocked but no admissible $p_t$ exists, the controller does not
invent a subgoal: it executes the active goal and defers recovery to
post-failure diagnosis. When an admissible $p_t$ is proposed, verification
withholds the blocked action and inserts the prerequisite first
(Section~\ref{subsec:ablation} ablates this module).

A single failure may be transient, so VBL permits up to $K-1$ local retries.
After $K$ consecutive failures on the same goal, a schema-constrained
diagnostic call returns
\[
  \textsc{Diagnose}(d(g_0),g_t,o_t,h_t,E_t)
  \rightarrow (z_t,p_t,e_t),
\]
where $o_t$ is the failing post-action observation, $z_t$ the attributed cause,
$p_t$ an optional prerequisite, and $e_t$ the evidence. Each invocation
prescribes one stack transition (Table~\ref{tab:attribution}).

\begin{table}[t]
\centering
\small
\caption{Failure attributions returned by \textsc{Diagnose} and the single
goal-stack transition each one prescribes.}
\label{tab:attribution}
\begin{adjustbox}{max width=\linewidth}
\begin{tabular}{lll}
\toprule
Attribution $z_t$ & Interpretation & State transition \\
\midrule
$\textsc{MissingPrecond}$ & prerequisite $p_t$ is absent
  & $\textsc{Push}(p_t)$ \\
$\textsc{WrongDecomp}$ & current decomposition is unsuitable
  & $\textsc{BacktrackTo}$ parent \\
$\textsc{Transient}$ & temporary execution failure
  & retry within $N_{\max}$ \\
$\textsc{Unrecoverable}$ & current path cannot progress
  & mark failed; $\textsc{BacktrackTo}$ parent \\
\bottomrule
\end{tabular}
\end{adjustbox}
\end{table}

If a proposed prerequisite is inadmissible, or a transient error exceeds
$N_{\max}$, the path is treated as incorrect or unrecoverable and the
controller backtracks to the parent. Thus \textsc{Verify} revises the active
goal before the next action when evidence already supports a stack update,
while \textsc{Diagnose} resumes planning after local retries still fail.

\subsection{Bounded Execution}
\label{subsec:bounded-execution}

The procedure uses a maximum stack depth $D_{\max}$, a per-goal retry limit
$N_{\max}$, and a global interaction budget $B$. Per-goal limits
$L_{\mathrm{push}}$ on pushes and $L_{\mathrm{bt}}$ on backtracks prevent
repeated decompositions that do not produce a new state.
These bounds also ensure termination: actions consume $B$, retries consume the
per-goal allowance, and stack changes consume the finite push and backtrack
allowances.

In the full VBL configuration reported in our experiments, completed and failed
trajectory segments are summarized into a short episode-local buffer
\[
  e_i=\langle d(g_i),\text{context},\text{action},\text{outcome},
  \text{attribution},\text{recovery}\rangle,
\]
with each field grounded in an environment observation or a recorded state
transition. Before execution, $\operatorname{Retrieve}(E_t,g_t)$ attaches
records associated with the active goal or its current failure type to $c_t$
when such records exist. The buffer is cleared when the task ends; cross-task
storage is not considered here. The buffer is an implementation detail for
limiting context rather than a separate memory contribution
(Section~\ref{sec:related}).

\paragraph{Reduction to ReAct.}
If verification and diagnosis are disabled, the stack remains $[g_0]$, the
episode buffer is unused ($\operatorname{Retrieve}$ returns nothing), and
$c_t$ reduces to the original instruction $d(g_0)=I$. Every action is then
generated by the base ReAct executor from the same interaction history and
action format.
Under this configuration, our ReAct baseline matches the executor with no
stack-conditioned control, enabling comparisons that vary only VBL's
additional control mechanisms. The ablations in Section~\ref{subsec:ablation}
sit between these two extremes: each disables exactly one of verification and
diagnosis while retaining the goal stack.
Algorithm~\ref{alg:vbl} is the control skeleton; the complete procedure is given in Appendix~\ref{app:alg}.

\section{Experiments}
\label{sec:experiments}

We evaluate whether structured precondition handling improves long-horizon
interactive agents relative to linear acting and goal-conditioned reflection.
Unless otherwise noted, all methods share the same backbone model, interaction
budget, and environment interface within each setting; the methods differ only
in their control layer, which for VBL comprises the goal stack, the verification
and diagnosis modules, and the episode-local buffer
(Section~\ref{subsec:bounded-execution}).

\subsection{Experimental Settings}
\label{subsec:setup}

\paragraph{Benchmarks.}
We evaluate on three complementary interactive environments that stress
long-horizon planning under latent preconditions:
\textbf{AppWorld}~\citep{trivedi2024appworldcontrollableworldapps} (interactive coding),
\textbf{ALFWorld}~\citep{shridhar2021alfworldaligningtextembodied} (text-embodied household),
and \textbf{ScienceWorld}~\citep{wang2022scienceworldagentsmarter5th} (scientific procedures).
We use the AppWorld \emph{Normal} ($168$) and \emph{Challenge} ($417$) test splits,
ALFWorld \texttt{valid\_unseen} ($134$), and ScienceWorld test ($211$ episodes / $24$ types);
detailed environment descriptions are in Appendix~\ref{app:benchmarks}.

\paragraph{Agent models.}
To assess robustness across model families, we instantiate every method with
three backbones spanning scale and access regime:
\textbf{Qwen3.5-9B} (open-weight, smaller-scale),
\textbf{GPT-4o} (closed-source, frontier-scale), and
\textbf{DeepSeek-V3.2} (open-weight, large-scale).
Holding the agent framework fixed while varying the backbone isolates the
contribution of VBL's control layer from raw model capacity.

\paragraph{Metrics.}
Our primary metric is task success rate (\%), i.e., the fraction of episodes for
which the terminal goal predicate $g^\star$ is satisfied within the interaction
budget (for AppWorld, an episode counts as successful only when all unit tests
pass). To characterize cost, we report average tokens per episode (kTok) and
tokens per solved task (kTok/S). Token counts include every LLM call uniformly
across benchmarks and methods---executor, \textsc{Verify}, \textsc{Diagnose},
and \textsc{Summarize} in the full VBL configuration---so controller overhead is
never omitted from the cost comparison.

\paragraph{Implementation details.}
All methods share the same environment adapters, ReAct action format, and
decoding configuration within each backbone.
We cap execution at $50$ environment steps on ScienceWorld, $50$ on ALFWorld,
and $40$/$60$ on AppWorld Normal/Challenge.
The per-episode LLM-call budget is $2B$ on every benchmark. For VBL we use a
maximum stack depth $D_{\max}$, a per-goal retry limit $N_{\max}$, per-goal
limits $L_{\mathrm{push}}$ and $L_{\mathrm{bt}}$, and a diagnosis threshold $K$.%
\footnote{Their values, together with the decoding configuration, are listed in
Appendix~\ref{app:impl}.}
Verification runs whenever $\mathit{due}$ is set
(Section~\ref{subsec:recovery}). Episode-local buffers are cleared when the
task ends. VBL is training-free; control bounds are shared across benchmarks,
while the step budget $B$ follows each environment's horizon.

\subsection{Baselines}
\label{subsec:baselines}

We compare VBL against three representative policies under a shared ReAct
executor and single-episode protocol (Section~\ref{sec:related}):
\textbf{NoThinking} emits actions directly from the observation and instruction
(lower-bound acting);
\textbf{ReAct}~\citep{yao2023react} interleaves free-form thoughts with
actions in a linear think--act--observe loop (the template VBL wraps---with
verification and diagnosis disabled, the executor matches this baseline;
Section~\ref{sec:method}); and
\textbf{ReflAct}~\citep{reflact2025} replaces next-action prediction with
goal-conditioned reflection, recovering via free-form textual revision rather
than structured stack operations.
No baseline maintains a comparable episode buffer, so ReAct--VBL isolates VBL's
control layer; we do not evaluate tree-search planners or multi-trial memory
agents.

\subsection{Main Results}
\label{subsec:results}

Table~\ref{tab:main_results} reports task success rates (\%) across benchmarks
and backbones. \textbf{VBL} is the best method on every backbone--benchmark pair.

\paragraph{Gains over the acting baseline.}
Relative to ReAct, the largest improvements appear where latent preconditions
dominate. On AppWorld \emph{Challenge}, VBL improves by up to $+22.07$ points
(Qwen3.5-9B, $8.63\!\rightarrow\!30.70$), with \emph{Normal}-split gains of
$+18.45$ points; the effect persists for stronger backbones ($+3.83$ and
$+5.95$ for DeepSeek-V3.2). The pattern extends to embodied and scientific
settings: up to $+10.45$ on ALFWorld (DeepSeek-V3.2) and $+14.21$ /
$+13.74$ on ScienceWorld (Qwen3.5-9B / DeepSeek-V3.2). Explicit
\textsc{Push}/\textsc{BacktrackTo} control recovers from unmet preconditions
more reliably than a linear think--act--observe loop.

\paragraph{Gains over reflection and consistency.}
Relative to ReflAct, structured stack operations remain more effective when the
failure root lies several steps upstream. ReflAct is not uniformly better than
ReAct---on some pairs it regresses (e.g., AppWorld with GPT-4o), consistent with
unconstrained reflection endorsing fluent but goal-inverting actions.
NoThinking is weakest on every backbone, confirming that VBL's advantage is not
merely the presence of any reasoning text, but the coupled
verification--diagnosis loop. Token cost is analyzed in
Section~\ref{subsec:efficiency}; per-type gains appear in
Appendix~\ref{app:per-type}.

\begin{table}[t]
\centering
\small
\caption{Main results: task success rate (\%) on AppWorld,
ALFWorld, and ScienceWorld. Chall.: AppWorld Challenge split; ALFW.: ALFWorld;
SciW.: ScienceWorld. VBL is the best method on every backbone--benchmark pair.}
\label{tab:main_results}
\setlength{\tabcolsep}{3pt}
\renewcommand{\arraystretch}{0.9}
\begin{tabular}{llcccc}
\toprule
& & \multicolumn{2}{c}{AppWorld} & & \\
\cmidrule(lr){3-4}
Model & Method & Normal & Chall. & ALFW. & SciW. \\
\midrule
\multirow{4}{*}{Qwen3.5-9B}
 & NoThinking & 14.29 & 7.43 & 69.40 & 21.80 \\
 & ReAct      & 17.26 & 8.63 & 81.34 & 33.18 \\
 & ReflAct    & 22.62 & 13.19 & 86.57 & 36.97 \\
 & VBL        & 35.71 & 30.70 & 87.31 & 47.39 \\
\midrule
\multirow{4}{*}{GPT-4o}
 & NoThinking & 17.86 & 10.55 & 67.16 & 35.55 \\
 & ReAct      & 48.81 & 30.22 & 85.07 & 50.71 \\
 & ReflAct    & 47.62 & 28.54 & 93.28 & 51.66 \\
 & VBL        & 50.00 & 31.65 & 94.78 & 57.35 \\
\midrule
\multirow{4}{*}{DeepSeek-V3.2}
 & NoThinking & 47.02 & 32.13 & 71.64 & 55.92 \\
 & ReAct      & 60.12 & 56.12 & 84.33 & 57.82 \\
 & ReflAct    & 61.31 & 48.68 & 86.57 & 66.35 \\
 & VBL        & 66.07 & 59.95 & 94.78 & 71.56 \\
\bottomrule
\end{tabular}
\end{table}

\subsection{Efficiency and Overhead}
\label{subsec:efficiency}

VBL adds verification and diagnosis calls, so a natural concern is overhead.
Because those modules are themselves LLM calls, we compare token usage with
all calls counted uniformly (Section~\ref{subsec:setup}).
Table~\ref{tab:efficiency} reports average tokens per episode (kTok) and tokens
per solved task (kTok/S\,$=100\times$\,kTok$/$SR) on Qwen3.5-9B.
Per episode, VBL consumes more tokens on every benchmark.
Per solved task, the cost--benefit picture is environment-dependent: on the
harder AppWorld splits the success gain lowers kTok/S relative to ReAct; on
easier ALFWorld the two methods are essentially tied; on ScienceWorld, despite
a large SR gain, kTok/S rises because controller overhead outpaces the success
improvement.

\begin{table}[t]
\centering
\small
\caption{Token cost on Qwen3.5-9B. AW-N/AW-C: AppWorld Normal/Challenge;
ALFW.: ALFWorld; SciW.: ScienceWorld. SR: success rate (\%); kTok: average
thousands of tokens per episode (including VBL's verification, diagnosis, and
summarization); kTok/S: $100\times$\,kTok$/$SR. \textbf{Bold}: higher SR or
lower kTok/S.}
\label{tab:efficiency}
\setlength{\tabcolsep}{5pt}
\renewcommand{\arraystretch}{1.15}
\begin{tabular}{@{}ll rrr@{}}
\toprule
Benchmark & Method & SR & kTok & kTok/S \\
\midrule
\multirow{2}{*}{AW-N}
 & ReAct & 17.26 & 138.9 & 804.8 \\
 & VBL   & \textbf{35.71} & 221.5 & \textbf{620.2} \\
\addlinespace[2pt]
\multirow{2}{*}{AW-C}
 & ReAct & 8.63 & 223.3 & 2587.5 \\
 & VBL   & \textbf{30.70} & 448.5 & \textbf{1460.9} \\
\addlinespace[2pt]
\multirow{2}{*}{ALFW.}
 & ReAct & 81.34 & 63.1 & 77.6 \\
 & VBL   & \textbf{87.31} & 67.1 & \textbf{76.9} \\
\addlinespace[2pt]
\multirow{2}{*}{SciW.}
 & ReAct & 33.18 & 66.6 & \textbf{200.7} \\
 & VBL   & \textbf{47.39} & 162.5 & 342.9 \\
\bottomrule
\end{tabular}
\end{table}

\subsection{Ablation Study}
\label{subsec:ablation}

To assess each control module, we compare full VBL against two leave-one-out
variants that disable exactly one module while retaining the goal stack and the
other:
\textbf{w/o \textsc{Verify}} removes activation and feedback-triggered checks,
and
\textbf{w/o \textsc{Diagnose}} removes post-failure attribution together with
the stack transitions and retry budget it prescribes.\footnote{Because
$N_{\max}$ is consumed only through the \textsc{Transient} branch of
\textsc{Diagnose}, removing \textsc{Diagnose} also removes the bounded
retry-and-rollback discipline. The drop should be read as the cost of losing
structured post-failure recovery as a whole, not attribution labels alone; an
untyped bounded fallback is left to future work.}
Table~\ref{tab:ablation} summarizes the results.

Full VBL outperforms both variants on every backbone--benchmark pair. Furthermore, the performance of some single-module variants is even lower than the ReAct baseline. The goal stack is therefore not a safe addition on its own; it pays off only when both modules govern it.

The two variants fail in characteristic ways. Without verification, a missing
prerequisite or misaligned goal is handled only once \textsc{Diagnose} fires
after $K$ consecutive failures. Without diagnosis, the verifier still inserts
prerequisites and backtracks on detected misalignment, but failures that
verification does not resolve are never attributed, so a goal judged aligned
and ready keeps spending actions at the wrong decomposition level.

\begin{table}[t]
\centering
\small
\caption{Leave-one-out ablation of VBL's two control modules across three
backbone models (Qwen3.5-9B, GPT-4o, DeepSeek-V3.2). Success rate (\%). Each
variant retains the goal stack and the other module.}
\label{tab:ablation}
\setlength{\tabcolsep}{4pt}
\renewcommand{\arraystretch}{1.05}
\begin{tabular}{l cc c c}
\toprule
& \multicolumn{2}{c}{AppWorld} & \multirow{2}{*}{ALFW.} & \multirow{2}{*}{SciW.} \\
\cmidrule(lr){2-3}
Variant & Normal & Chall. & & \\
\midrule
\multicolumn{5}{l}{\textit{Qwen3.5-9B}} \\
\quad VBL (full)             & 35.71 & 30.70 & 87.31 & 47.39 \\
\quad w/o \textsc{Verify}    & 35.12 & 28.06 & 76.12 & 40.28 \\
\quad w/o \textsc{Diagnose}  & 34.52 & 28.54 & 79.85 & 41.23 \\
\midrule
\multicolumn{5}{l}{\textit{GPT-4o}} \\
\quad VBL (full)             & 50.00 & 31.65 & 94.78 & 57.35 \\
\quad w/o \textsc{Verify}    & 44.05 & 29.50 & 93.28 & 53.08 \\
\quad w/o \textsc{Diagnose}  & 39.29 & 26.62 & 82.84 & 48.34 \\
\midrule
\multicolumn{5}{l}{\textit{DeepSeek-V3.2}} \\
\quad VBL (full)             & 66.07 & 59.95 & 94.78 & 71.56 \\
\quad w/o \textsc{Verify}    & 61.90 & 52.52 & 87.31 & 62.56 \\
\quad w/o \textsc{Diagnose}  & 64.88 & 51.56 & 81.34 & 69.19 \\
\bottomrule
\end{tabular}
\end{table}

\subsection{Recovery-Operation Analysis}
\label{subsec:recovery-analysis}

Beyond aggregate success, we ask \emph{which} stack operations VBL actually uses
on AppWorld, where multi-app control flow makes latent preconditions most
severe. Table~\ref{tab:control-ops} reports \textsc{Push} and
\textsc{BacktrackTo} frequencies.

\textsc{Push} is the dominant remedy, firing in $33.9$--$93.5\%$ of episodes:
most unmet preconditions are resolved by inserting an enabling subgoal rather
than discarding the plan. \textsc{BacktrackTo} is less frequent, but its rate
tracks wrong-subgoal risk---on every backbone, Challenge episodes trigger
backtracking more often than Normal ones (Qwen3.5-9B:
$13.7\%\!\rightarrow\!27.3\%$; GPT-4o: $4.8\%\!\rightarrow\!8.2\%$;
DeepSeek-V3.2: $4.0\%\!\rightarrow\!16.8\%$).
\textsc{Push} need not rise in parallel: for DeepSeek-V3.2 it is marginally
lower on Challenge ($65.2\%$ vs.\ $69.0\%$) even as backtracking
quadruples, consistent with harder episodes failing more often from a wrong
decomposition than from a nameable missing prerequisite.
Together with the ablation, this supports a coupled design: a high-frequency
precondition inserter paired with a difficulty-sensitive stack repair.

\begin{table}[t]
\centering
\small
\caption{Control-operation usage for VBL on AppWorld. AW-N/AW-C: AppWorld
Normal/Challenge. ``mean'' is the average number of operations per episode;
``\%ep'' is the fraction of episodes with at least one such operation.
\textsc{Push} remains the dominant remedy; \textsc{BacktrackTo} is secondary but
consistently higher on the harder Challenge split.}
\label{tab:control-ops}
\setlength{\tabcolsep}{4pt}
\renewcommand{\arraystretch}{1.0}
\begin{tabular}{ll cc cc}
\toprule
& & \multicolumn{2}{c}{Push} & \multicolumn{2}{c}{Backtrack} \\
\cmidrule(lr){3-4}\cmidrule(lr){5-6}
Bench. & Model & mean & \%ep & mean & \%ep \\
\midrule
\multirow{3}{*}{AW-N}
 & Qwen3.5-9B    & 1.10 & 63.1 & 0.22 & 13.7 \\
 & GPT-4o        & 0.42 & 33.9 & 0.05 & 4.8 \\
 & DeepSeek-V3.2 & 0.77 & 69.0 & 0.04 & 4.0 \\
\midrule
\multirow{3}{*}{AW-C}
 & Qwen3.5-9B    & 1.55 & 93.5 & 0.47 & 27.3 \\
 & GPT-4o        & 0.89 & 71.2 & 0.12 & 8.2 \\
 & DeepSeek-V3.2 & 0.77 & 65.2 & 0.18 & 16.8 \\
\bottomrule
\end{tabular}
\end{table}

\subsection{Case Study}
\label{subsec:case-studies}

To illustrate \emph{how} VBL converts unmet preconditions into successful trajectories, we examine an AppWorld episode using Qwen3.5-9B. In this episode, both ReAct and ReflAct fail, whereas VBL succeeds through a verification-triggered \textsc{Push}. A complementary example involving diagnosis and backtracking is provided in Appendix~\ref{app:case-diagnose}. The task requires settling shared dinner expenses through Venmo: identifying the roommates, sending payment requests for the taxi fare, and paying Nancy for the food.\footnote{Task \texttt{2d9f728\_1}. VBL passes $10/10$ tests with one \textsc{Push}, compared with $1/10$ for ReAct and $4/10$ for ReflAct.}

Protected phone and Venmo endpoints return HTTP 401 when called without an access token. ReAct discovers this requirement only after a failed call. It then oscillates among unauthorized requests, login attempts with incorrect argument formats, and unproductive API-documentation lookups until reaching the step limit. ReflAct retains the overall objective but still repeatedly issues unauthorized calls or malformed search requests, failing to establish authentication reliably.

VBL, by contrast, treats authentication as an explicit precondition on the goal stack. The first 401 response raises $\mathit{due}$, causing \textsc{Verify} to run before the next protected call. \textsc{Verify} determines that the current goal remains aligned with the task but is \emph{blocked} by missing authentication, and returns this prerequisite to \textsc{Push} as an enabling subgoal. Crucially, this intervention occurs without waiting for the $K$ consecutive failures required to trigger \textsc{Diagnose}. Once the prerequisite has been satisfied, the agent obtains the required tokens before invoking the payment APIs, and the remaining calls proceed successfully (Figure~\ref{fig:case-verify}).

The underlying ReAct loop remains unchanged; the improvement comes from inserting the missing enabling goal onto the stack, thereby turning a cascade of 401 errors into a short authenticated trajectory. This episode illustrates the feedback-triggered branch of verification described in Section~\ref{sec:method}.

\section{Conclusion}
\label{sec:conclusion}

We presented the \textbf{Verify--Backtrack Loop (VBL)}, a training-free
extension of ReAct that maintains task decomposition as a bounded goal stack
and converts verification and typed failure attribution into structured
control operations (\textsc{Push}, \textsc{Pop}, \textsc{BacktrackTo}). Across
three interactive benchmarks and three backbones, VBL improves task success
over linear acting and reflection-based control, with the largest gains where
latent preconditions dominate; ablations show that removing either verification
or diagnosis degrades performance, in several settings to below the ReAct
baseline, so the two modules are complementary rather than redundant.

Several limitations remain. Both control modules rely on the backbone's own
judgment, so mistaken readiness or attribution decisions can corrupt the
stack. Because backtracking revises planning state without reversing executed
actions, VBL cannot undo irreversible side effects. The control layer also
costs additional tokens. Natural next steps are stronger verification and
diagnosis, true rollback when actions are reversible, and leaner control that
preserves the same stack discipline at lower overhead.

\begin{figure}[!h]
\centering
\footnotesize
\setlength{\tabcolsep}{4pt}
\begin{tabular}{@{}p{0.95\columnwidth}@{}}
\toprule
\textbf{Task:} settle shared dinner on Venmo (taxi requests + pay Nancy) \\
\midrule
\textbf{ReAct / ReflAct (fail).} \texttt{search\_contacts}/\texttt{search\_users} $\rightarrow$ \textbf{401} (no token) $\rightarrow$ retry unauthorized / ill-formed calls $\rightarrow$ step limit or incomplete settlement. \hfill ReAct $1/10$,\ ReflAct $4/10$ \\[0.4em]
\textbf{VBL (succeed).}\\[0.2em]
\begin{tabular}{@{}r@{\,}p{0.82\columnwidth}@{}}
1. & \texttt{phone.search\_contacts(``Nancy'')} $\rightarrow$ \textbf{401} (token missing) \\[0.1em]
2. & \fbox{\textsc{Verify}$\rightarrow$\textsc{Push}(login)}: \texttt{phone.login}/\texttt{venmo.login} $\rightarrow$ tokens issued \\[0.1em]
3. & \texttt{create\_payment\_request}(``Taxi''), \texttt{create\_transaction}(``Food'') $\rightarrow$ sent. \hfill \textbf{$10/10$} \\
\end{tabular} \\
\bottomrule
\end{tabular}
\caption{Case study (task \texttt{2d9f728\_1}). Baselines thrash on auth errors; VBL inserts login via \textsc{Push} and completes the settlement.}
\label{fig:case-verify}
\end{figure}

\FloatBarrier

\bibliographystyle{unsrtnat}
\bibliography{references}

\clearpage
\appendix
\counterwithin{table}{section}
\counterwithin{figure}{section}
\counterwithin{algorithm}{section}

\begin{center}
{\LARGE\bfseries Appendix}
\end{center}
\vspace{1em}

\section{Benchmark Details}
\label{app:benchmarks}

This section expands the three environments introduced in the experimental
setup of Section~\ref{subsec:setup}.

\paragraph{AppWorld.}
AppWorld~\citep{trivedi2024appworldcontrollableworldapps} is a high-fidelity
interactive coding benchmark in which agents complete day-to-day digital tasks
by writing executable code against a suite of simulated applications and APIs.
Tasks typically require multi-app control flow, authentication, identifier
resolution, and recovery from API failures, making unmet preconditions a
dominant failure mode. We evaluate on the \emph{Normal} ($168$ tasks) and
\emph{Challenge} ($417$ tasks) test splits.

\paragraph{ALFWorld.}
ALFWorld~\citep{shridhar2021alfworldaligningtextembodied} aligns text-based
household goals with embodied ALFRED-style tasks. Agents navigate rooms,
manipulate objects, and satisfy compositional instructions through
text actions, so success depends on sequencing enabling subgoals (e.g.,
locating, cleaning, or heating an object) before the terminal action.
We evaluate on the \texttt{valid\_unseen} split ($134$ episodes).

\paragraph{ScienceWorld.}
ScienceWorld~\citep{wang2022scienceworldagentsmarter5th} is a text
environment for elementary scientific experiments. Agents must operate
instruments, track changing world state, and complete multi-step procedures,
again exposing the need to establish experimental preconditions rather than
jumping to the final measurement or conclusion. We evaluate on the test split
($211$ episodes spanning $24$ task types).

Together, the three benchmarks cover interactive coding, household embodiment,
and scientific procedural reasoning, providing a broad testbed for structured
precondition recovery.

\section{Full VBL Procedure}
\label{app:alg}

Section~\ref{sec:method} states the control skeleton of VBL.
Algorithm~\ref{alg:vbl-full} expands the procedure used in all experiments,
including the $\mathit{due}$ gate, admissibility constraints, episode-buffer
updates, and bounded push/backtrack allowances. Hyperparameters for this
procedure are listed in Appendix~\ref{app:impl}.
A prerequisite $p$ is \emph{admissible} for $g$ iff it is concrete and
actionable, $p\notin P(g)$ and $p\notin S$, the resulting depth stays within
$D_{\max}$, and $g$ has fewer than $L_{\mathrm{push}}$ pushes; each
$\textsc{BacktrackTo}$ consumes one of the $L_{\mathrm{bt}}$ backtracks of its
target, and is skipped once that allowance is exhausted.

\begin{algorithm}[!t]
\caption{Verify--Backtrack Loop (full procedure)}
\label{alg:vbl-full}
\small
\begin{algorithmic}[1]
\Require environment $\mathcal{M}$, ReAct policy $\pi_\theta$, bounds
$D_{\max},N_{\max},L_{\mathrm{push}},L_{\mathrm{bt}},B$, threshold $K$
\State $S\gets[g_0]$ with $d(g_0)\gets I$;
$E\gets\varnothing$; $t\gets0$; $\mathit{due}\gets\mathrm{true}$
\State initialize history $h$ from $(I,o_0)$
\While{$S\neq\varnothing$ \textbf{and} $t<B$}
  \State $g\gets\mathrm{top}(S)$; $o\gets$ current observation
  \If{$\mathit{due}$}
    \State $\mathit{due}\gets\mathrm{false}$
    \State $(q,u,p,e)\gets\textsc{Verify}(d(g_0),g,o,h,E)$
    \Comment{$d(g_0)=I$ only}
    \If{$q=\mathrm{complete}$}
      \State $E\gets\textsc{Summarize}(E,g,h)$;
      $\textsc{Pop}()$; $\mathit{due}\gets\mathrm{true}$; \textbf{continue}
    \ElsIf{$q=\mathrm{misaligned}$}
      \State $E\gets\textsc{Summarize}(E,g,h)$;
      $\textsc{BacktrackTo}$ parent;
      $\mathit{due}\gets\mathrm{true}$; \textbf{continue}
    \ElsIf{$u=\mathrm{blocked}$ \textbf{and} $p$ is admissible}
      \State $P(g)\gets P(g)\cup\{p\}$; $\textsc{Push}(p)$;
      $\mathit{due}\gets\mathrm{true}$; \textbf{continue}
    \EndIf
  \EndIf
  \State $c\gets(d(g_0),d(g),\operatorname{Retrieve}(E,g))$
  \State $a\sim\pi_\theta(\cdot\mid h,c)$; execute $a$; observe $o'$;
  append $(a,o')$ to $h$; $t\gets t+1$
  \If{$g$ is complete under $o'$}
    \State $E\gets\textsc{Summarize}(E,g,h)$;
    $\textsc{Pop}()$; $\mathit{due}\gets\mathrm{true}$; \textbf{continue}
  \EndIf
  \If{$o'$ indicates failure}
    \State $n_{\mathrm{err}}(g)\gets n_{\mathrm{err}}(g)+1$;
    $\mathit{due}\gets\mathrm{true}$
    \Comment{re-verify before the next action}
    \If{$n_{\mathrm{err}}(g)\geq K$}
      \State $(z,p,e)\gets\textsc{Diagnose}(d(g_0),g,o',h,E)$
      \State $E\gets\textsc{Update}(E,g,a,o',z,p,e)$;
      $n_{\mathrm{err}}(g)\gets0$
      \If{$z=\textsc{MissingPrecond}$ \textbf{and} $p$ admissible}
        \State $P(g)\gets P(g)\cup\{p\}$; $\textsc{Push}(p)$;
        $\mathit{due}\gets\mathrm{true}$
      \ElsIf{$z=\textsc{Transient}$ \textbf{and} $n(g)<N_{\max}$}
        \State $n(g)\gets n(g)+1$
      \Else
        \State mark path failed; $\textsc{BacktrackTo}$ parent;
        $\mathit{due}\gets\mathrm{true}$
      \EndIf
    \EndIf
  \ElsIf{$o'$ indicates a material state change}
    \State $\mathit{due}\gets\mathrm{true}$
  \EndIf
\EndWhile
\State \Return success and interaction count $t$; discard $E$
\end{algorithmic}
\end{algorithm}

\section{Implementation Details}
\label{app:impl}

Table~\ref{tab:hyperparams} lists the single shared control configuration used
for full VBL in Algorithm~\ref{alg:vbl-full}, together with the interaction
budgets stated in Section~\ref{subsec:setup}. Control bounds
$(D_{\max},N_{\max},L_{\mathrm{push}},L_{\mathrm{bt}},K)$ are held fixed across
benchmarks; only the environment step budget $B$ follows each benchmark's
horizon. All LLM calls---executor, \textsc{Verify}, \textsc{Diagnose}, and
\textsc{Summarize}---use greedy decoding (temperature $0$, no nucleus
truncation), so a run is deterministic up to provider-side non-determinism, and
all methods share this decoding configuration within a backbone. Reported token
costs count every such call the same way on all benchmarks; no controller call
is excluded from kTok.

\begin{table}[htbp]
\centering
\small
\caption{Hyperparameters for full VBL. Control bounds are shared across
benchmarks; $B$ matches the experimental setting in Section~\ref{subsec:setup}.
$^{a}$Normal / Challenge.}
\label{tab:hyperparams}
\setlength{\tabcolsep}{4pt}
\renewcommand{\arraystretch}{1.0}
\begin{tabular}{@{}l ccc@{}}
\toprule
& AppWorld & ALFWorld & ScienceWorld \\
\midrule
Stack depth $D_{\max}$        & 6 & 6 & 6 \\
Retry limit $N_{\max}$        & 8 & 8 & 8 \\
Push limit $L_{\mathrm{push}}$ & 4 & 4 & 4 \\
\quad at the root goal        & 1 & 1 & 1 \\
Backtrack limit $L_{\mathrm{bt}}$ & 3 & 3 & 3 \\
Diagnosis threshold $K$       & 2 & 2 & 2 \\
Env.\ step budget $B$         & 40 / 60$^{a}$ & 50 & 50 \\
LLM-call budget               & $2B$ & $2B$ & $2B$ \\
Temperature                   & 0 & 0 & 0 \\
\bottomrule
\end{tabular}
\end{table}

\section{Control-Module Output Schemas}
\label{app:schemas}

\textsc{Verify} and \textsc{Diagnose} are schema-constrained side-channel calls.
They are never appended into the executor's think--act dialogue as free-form
thoughts. Below we give the paper-level fields used by
Algorithm~\ref{alg:vbl-full} and the concrete JSON shapes used in experiments.
Field names in JSON may differ slightly by environment wording, but the
semantics match Section~\ref{subsec:recovery}.

\paragraph{Verify.}
A call returns
\[
  \textsc{Verify}(d(g_0),g_t,o^{\mathrm{curr}}_t,h_t,E_t)
  \rightarrow (q_t,u_t,p_t,e_t),
\]
with alignment
$q_t\in\{\mathrm{aligned},\mathrm{misaligned},\mathrm{complete}\}$,
readiness
$u_t\in\{\mathrm{ready},\mathrm{blocked},\mathrm{unknown}\}$, an optional
prerequisite $p_t$, and brief evidence $e_t$. The experimental JSON is:
\promptbox{%
\{``alignment'': ``aligned'' $|$ ``misaligned'' $|$ ``complete'',\\
 ``readiness'': ``ready'' $|$ ``blocked'' $|$ ``unknown'',\\
 ``prerequisite'': ``$<$one concrete subgoal if blocked, else empty$>$'',\\
 ``evidence'': ``$<$brief$>$''\}
}
When readiness is \texttt{blocked} and $p_t$ is admissible, the controller
issues $\textsc{Push}(p_t)$. When alignment is \texttt{misaligned}, it issues
$\textsc{BacktrackTo}$ the parent. When alignment is \texttt{complete}, it
issues $\textsc{Pop}$. If readiness is \texttt{unknown}, or the goal is blocked
but no admissible $p_t$ exists, the controller executes the active goal and
defers recovery to post-failure diagnosis---matching Section~\ref{subsec:recovery}.

In the released controller, the same decision is often emitted in a compact
form that the router expands into $(q_t,u_t,p_t,e_t)$:
\promptbox{%
\{``ready'': 1 or 0,\\
 ``missing\_precondition'': ``$<$subgoal if ready=0, else empty$>$'',\\
 ``reason'': ``$<$brief$>$''\}
}
Here \texttt{ready=1} maps to $u_t=\mathrm{ready}$; \texttt{ready=0} with a
concrete precondition maps to $u_t=\mathrm{blocked}$ and $p_t$; completion and
misalignment are routed by the controller when the observation or stack state
supports $\textsc{Pop}$ / $\textsc{BacktrackTo}$.

\paragraph{Diagnose.}
After $K$ consecutive failures on the same goal, diagnosis returns
\[
  \textsc{Diagnose}(d(g_0),g_t,o_t,h_t,E_t)
  \rightarrow (z_t,p_t,e_t),
\]
where $z_t$ is one of the four attributions in Table~\ref{tab:attribution}
(\textsc{MissingPrecond}, \textsc{WrongDecomp}, \textsc{Transient},
\textsc{Unrecoverable}). The experimental JSON is:
\promptbox{%
\{``type'': ``MISSING\_PRECOND'' $|$ ``WRONG\_DECOMPOSITION''\\
\hspace*{2.2em}$|$ ``TRANSIENT'' $|$ ``UNRECOVERABLE'',\\
 ``precondition'': ``$<$subgoal if MISSING\_PRECOND, else empty$>$'',\\
 ``reason'': ``$<$brief$>$''\}
}
The mapping to stack transitions is exactly the attribution table
(Table~\ref{tab:attribution}):
\textsc{MissingPrecond}$\rightarrow$\textsc{Push},
\textsc{WrongDecomp}/\textsc{Unrecoverable}$\rightarrow$\textsc{BacktrackTo}
parent, and \textsc{Transient}$\rightarrow$ local retry within $N_{\max}$.
Vague proposals (e.g., ``explore'', ``read docs'') are rejected; the call is
treated as \textsc{Transient} so the executor can retry under the existing
action format.

\section{Episode Buffer and Retrieve}
\label{app:buffer}

Full VBL maintains an episode-local buffer $E_t$ that stores short grounded
records of completed or failed trajectory segments. As in
Section~\ref{subsec:bounded-execution}, each record has the form
\begin{multline*}
  e_i=\langle d(g_i),\;\mathrm{context},\;\mathrm{action},\\
  \mathrm{outcome},\;\mathrm{attribution},\;\mathrm{recovery}\rangle.
\end{multline*}
Concrete fields used in the implementation are listed in
Table~\ref{tab:buffer-fields}. Every field is grounded in an environment
observation or a recorded stack transition; the buffer does not store
cross-task memory.

\begin{table}[htbp]
\centering
\small
\caption{Episode-buffer record fields. Records are written by
\textsc{Summarize}/\textsc{Update} when a goal is popped, a path is marked
failed, or diagnosis assigns a recovery transition.}
\label{tab:buffer-fields}
\setlength{\tabcolsep}{3pt}
\renewcommand{\arraystretch}{1.05}
\begin{tabular}{@{}lp{0.72\linewidth}@{}}
\toprule
Field & Content \\
\midrule
$d(g_i)$ & Active-goal description at the time of the segment \\
context & Short local context (recent observation / stack focus) \\
action & Action (or code block) that closed the segment \\
outcome & Post-action observation or success/failure signal \\
attribution & Diagnose label $z$ when the segment ended in failure;
  empty on clean $\textsc{Pop}$ \\
recovery & Stack operation applied
  ($\textsc{Push}$ / $\textsc{Pop}$ / $\textsc{BacktrackTo}$ / retry) \\
\bottomrule
\end{tabular}
\end{table}

\paragraph{Write rules.}
\textsc{Summarize} appends a record when a goal is completed and popped, or when
a path is abandoned. \textsc{Update} appends or refreshes a record when
\textsc{Diagnose} fires, storing the failing action, observation, attribution
$z_t$, optional prerequisite $p_t$, and the prescribed recovery. The buffer is
cleared when the episode ends.

\paragraph{Retrieve rules.}
Before the executor samples $a_t$, the controller forms
\[
  c_t=\bigl(d(g_0),\,d(g_t),\,\operatorname{Retrieve}(E_t,g_t)\bigr).
\]
$\operatorname{Retrieve}(E_t,g_t)$ returns at most a few records that match
either (i)~the active goal description $d(g_t)$ (same or closely related
subgoal text) or (ii)~the current failure type when the latest segment on
$g_t$ carries an attribution. Records for other stack nodes are withheld.
If no matching record exists, $\operatorname{Retrieve}$ returns the empty
string and $c_t$ reduces to $(d(g_0),d(g_t))$. Disabling verification and
diagnosis leaves the stack at $[g_0]$ and makes $\operatorname{Retrieve}$ a
no-op, recovering the ReAct baseline as stated in Section~\ref{sec:method}.

\section{Prompts and ICL Examples}
\label{app:prompts}

All methods share the same environment adapter and action interface within each
benchmark. They differ only in the control text shown to the model: turn format,
optional one-shot ICL, and---for VBL---side-channel schemas from
Appendix~\ref{app:schemas}. VBL's executor uses the ReAct format; stack updates
change $d(g_t)$ (and retrieved buffer records) without rewriting the ReAct
instruction.

\subsection{Shared turn formats}
\label{app:turn-formats}

\begin{itemize}
  \item \textbf{NoThinking:} emit only the environment action (no Thought /
  Reflection).
  \item \textbf{ReAct / VBL executor:}
  \texttt{Thought:} $\ldots$ then one action.
  \item \textbf{ReflAct:}
  \texttt{Reflection:} one sentence on current state relative to the task goal,
  then one action.
\end{itemize}

\subsection{AppWorld}
\label{app:prompts-appworld}

Agents interact by writing one fenced Python code block per turn that calls
\texttt{apis.<app\_name>.<api\_name>(...)}. The shared protocol prompt covers API discovery
(\texttt{show\_app\_descriptions} / \texttt{show\_api\_descriptions} /
\texttt{show\_api\_doc}), supervisor credentials, authentication via
\texttt{access\_token}, and finishing with
\texttt{apis.supervisor.complete\_task(...)}. Method-specific turn formats and
the compact one-shot ICL used for baselines are:

\paragraph{NoThinking format + ICL (excerpt).}
\promptbox{%
NoThinking output format: write exactly ONE fenced Python code block\\
(no Thought / Reflection).\\[0.35em]
Example task: Pay my roommate \$20 on Venmo for last week's groceries.\\[0.2em]
```python\\
passwords = apis.supervisor.show\_account\_passwords()\\
print(passwords)\\
print(apis.api\_docs.show\_api\_descriptions(\\
\hspace*{2.2em}app\_name='venmo'))\\
```\\
Observation: \{...passwords...\} \{...venmo API list...\}\\[0.2em]
```python\\
profile = apis.supervisor.show\_profile()\\
result = apis.venmo.login(\\
\hspace*{2.2em}username=profile[``email''],\\
\hspace*{2.2em}password=passwords[``venmo''])\\
print(result)\\
```
}

\paragraph{ReAct format + ICL (excerpt).}
\promptbox{%
Thought: $<$reasoning$>$\\
```python\\
\# one code action\\
```\\[0.35em]
Example (same Venmo payment task):\\
Thought: I need supervisor credentials and Venmo API docs before logging in.\\
```python\\
passwords = apis.supervisor.show\_account\_passwords()\\
print(passwords)\\
print(apis.api\_docs.show\_api\_descriptions(\\
\hspace*{2.2em}app\_name='venmo'))\\
```\\
Observation: \{...\}\\
Thought: I have credentials. Next I log into Venmo to obtain an access\_token.\\
```python\\
profile = apis.supervisor.show\_profile()\\
result = apis.venmo.login(\\
\hspace*{2.2em}username=profile[``email''],\\
\hspace*{2.2em}password=passwords[``venmo''])\\
print(result)\\
```
}

\paragraph{ReflAct format + ICL (excerpt).}
\promptbox{%
Reflection: $<$one sentence on state relative to the task goal$>$\\
```python\\
\# one code action\\
```\\[0.35em]
Example (same Venmo payment task):\\
Reflection: I have not yet authenticated to Venmo or identified the\\
roommate, so I am still at the start of completing the \$20 payment.\\
```python\\
passwords = apis.supervisor.show\_account\_passwords()\\
print(passwords)\\
print(apis.api\_docs.show\_api\_descriptions(\\
\hspace*{2.2em}app\_name='venmo'))\\
```\\
Observation: \{...\}\\
Reflection: I now have credentials but still no Venmo access\_token,\\
which I need before I can send the payment.\\
```python\\
profile = apis.supervisor.show\_profile()\\
result = apis.venmo.login(...)\\
print(result)\\
```
}

\paragraph{VBL side-channel prompts (AppWorld).}
The executor prompt is the ReAct protocol above. \textsc{Verify} /
\textsc{Diagnose} receive the root task text, the goal-stack view (top =
current focus), a short recent-execution window, and the schema instructions
from Appendix~\ref{app:schemas}. Representative instruction text:
\promptbox{%
Verify (non-root): Before executing the TOP goal, decide whether it can\\
be attempted now. ready=0 ONLY for a specific missing prerequisite\\
(e.g.\ must log in, must fetch an id). Never propose vague explore /\\
read-docs steps. missing\_precondition must be concrete and different\\
from the TOP goal.\\[0.35em]
Verify (root): check ONLY for a missing login/auth step; do not\\
decompose the root into other subgoals.\\[0.35em]
Diagnose: The TOP goal failed twice in a row. Classify as\\
MISSING\_PRECOND / WRONG\_DECOMPOSITION / TRANSIENT / UNRECOVERABLE.\\
precondition must be a concrete state to obtain; vague API-discovery\\
proposals are treated as TRANSIENT.
}

\subsection{ALFWorld}
\label{app:prompts-alfworld}

Instructions follow the ReflAct Appendix~K.1 household template. The agent
emits \texttt{Thought:}/\texttt{Reflection:} plus \texttt{Action:} from the
admissible command set (\texttt{go to}, \texttt{take}, \texttt{put},
\texttt{open}/\texttt{close}, \texttt{use}, \texttt{clean}/\texttt{heat}/
\texttt{cool}). Each method uses one-shot ICL keyed by task type
(\texttt{pick\_and\_place}, \texttt{pick\_clean\_then\_place},
\texttt{pick\_heat\_then\_place}, \texttt{pick\_cool\_then\_place},
\texttt{pick\_two\_obj}, \texttt{look\_at\_obj}). NoThinking reuses the ReAct
ICL with Thought lines stripped; ReflAct uses Reflection-rewritten
trajectories; VBL's executor shares the ReAct instruction and ICL bank.

\paragraph{ReAct instruction (excerpt).}
\promptbox{%
Interact with a household to solve a task. ... For each turn, think about\\
the current condition and plan future actions, then output:\\
``Thought: ... / Action: ...'' (one action per turn).\\
Available actions: go to recep; take obj from recep; put obj in/on recep;\\
open/close recep; use obj; clean/heat/cool obj with recep.
}

\paragraph{One-shot ICL excerpt} (\texttt{pick\_and\_place}; ReAct):
\promptbox{%
Task: put some spraybottle on toilet.\\
Thought: To solve the task, I need to find and take a spraybottle, then\\
put it on toilet. ... starting with cabinet 1.\\
Action: go to cabinet 1\\
Observation: On the cabinet 1, you see a cloth 1, a soapbar 1, ...\\
Action: go to cabinet 2\\
Observation: The cabinet 2 is closed.\\
Action: open cabinet 2\\
Observation: ... spraybottle 2.\\
Thought: Now I find a spraybottle (2). Next, I need to take it.\\
Action: take spraybottle 2 from cabinet 2\\
...\\
Action: put spraybottle 2 in/on toilet 1
}

\paragraph{ReflAct turn format.}
Same admissible actions; each turn is
\texttt{Reflection:} (state relative to the goal) then
\texttt{Action:} (one command).
The corresponding ICL rewrites the trajectory above with state-grounded
reflections (e.g., ``Currently, I am at cabinet~2 and have found a
spraybottle~2\ldots'').

\paragraph{VBL side channels.}
\textsc{Verify}/\textsc{Diagnose} use the same JSON schemas as
Appendix~\ref{app:schemas}, with household-specific rules: prerequisites must
be concrete admissible commands (e.g., \texttt{go to desk 1},
\texttt{open fridge 1}); vague ``search everywhere'' proposals are rejected.
When a non-root goal is active, the executor observation is augmented with a
short focus line of the form
``Current subgoal focus: $d(g_t)$''.

\subsection{ScienceWorld}
\label{app:prompts-sciworld}

Instructions follow the ReflAct Appendix~K.2 laboratory template (rooms,
teleport, focus, mix, etc.). One-shot ICL is the green-paint mixing trajectory
from that appendix; ReflAct uses the Reflection-rewritten variant; NoThinking
strips Thought/Reflection lines; VBL shares the ReAct instruction and ICL.

\paragraph{ReAct instruction (excerpt).}
\promptbox{%
You are a helpful assistant to do a scientific experiment. ...\\
Output format: ``Thought: ... / Action: ...'' (one action per turn).\\
Available actions include: open/close, activate/deactivate, connect,\\
use, look around, examine, pour, mix, teleport to LOC, focus on OBJ, wait.
}

\paragraph{One-shot ICL excerpt} (create green paint; ReAct):
\promptbox{%
Task: use chemistry to create green paint. When done, focus on the green paint.\\
Thought: Materials are likely in an art studio.\\
Action: look around\\
Observation: This room is called the bedroom. ...\\
Thought: The bedroom doesn't have the materials. Go to the art studio.\\
Action: teleport to art studio\\
...\\
Action: pour cup containing blue paint ... in cup containing nothing\\
Action: pour cup containing yellow paint ... in cup containing blue paint\\
Action: mix cup containing blue paint and yellow paint\\
Observation: Blue and yellow paint mix to produce green paint.\\
Action: focus on green paint
}

\paragraph{ReflAct / NoThinking.}
ReflAct replaces Thought with a state-relative Reflection (location, inventory,
focused object). NoThinking emits only \texttt{Action: ...}. VBL side-channel
prompts reuse the Appendix~\ref{app:schemas} JSON schemas with
ScienceWorld-specific admissible-action constraints.

\section{Model and Inference Settings}
\label{app:models}

\paragraph{Backbones.}
Every method is evaluated under three backbones, held fixed within each
comparison:
\begin{itemize}
  \item \textbf{Qwen3.5-9B} --- open-weight; served locally via an
  OpenAI-compatible vLLM endpoint (\texttt{--served-model-name Qwen3.5-9B}).
  \item \textbf{GPT-4o} --- closed-source Chat Completions API.
  \item \textbf{DeepSeek-V3.2} --- open-weight large model via API (or an
  OpenAI-compatible endpoint with served name \texttt{deepseek-v3.2}).
\end{itemize}

\paragraph{Decoding and budgets.}
All calls use temperature $0$ (greedy). Nucleus/top-$p$ sampling is disabled.
Within a backbone, NoThinking, ReAct, ReflAct, and VBL share the same client
configuration. Per-episode environment steps use the $B$ values in
Table~\ref{tab:hyperparams}; the LLM-call budget is $2B$ and counts executor
and controller calls uniformly. Approximate completion-token caps used by the
controller are: executor $512$--$1536$ (environment-dependent),
\textsc{Verify} $200$, \textsc{Diagnose} $300$, and short
\textsc{Summarize}/\textsc{Update} writes. Trajectories are truncated when the
serialized history exceeds a fixed character budget (AppWorld default
$20{,}000$ characters), matching the shared ReAct harness.

\paragraph{Token accounting.}
Reported kTok sums prompt and completion tokens over every LLM call in the
episode, including VBL's \textsc{Verify}, \textsc{Diagnose}, and
\textsc{Summarize}. No controller call is omitted from the cost tables in
Table~\ref{tab:efficiency}.

\section{Additional Case Study: Diagnostic Backtracking}
\label{app:case-diagnose}

Section~\ref{subsec:case-studies} illustrates verification-time \textsc{Push}. Here we complement
that example with diagnostic recovery on another AppWorld normal-split episode
with Qwen3.5-9B, where free-form reflection commits to a goal-inverting
operator while VBL repairs the plan via \textsc{BacktrackTo}.

A task asks the agent to \emph{request} \$13 publicly from Stacy on Venmo with
note ``For yesterday's meal.''\footnote{Task \texttt{024c982\_1}. VBL passes
$7/7$ tests ($1$ push, $1$ backtrack); ReAct $6/7$; ReflAct $1/7$.}
The critical distinction is operator choice:
\texttt{create\_payment\_request} vs.\ \texttt{create\_transaction}. ReflAct
correctly locates Stacy after some auth noise, but then reflects itself into
sending money (\texttt{create\_transaction})---a coherent yet goal-inverting
action that mutates \texttt{venmo.Transaction} instead of creating a payment
request, so nearly all unit tests fail. ReAct issues the right request API but
still fails the answer check, illustrating that linear recovery can get close
without reliable goal--operator alignment.

VBL keeps ``request payment from Stacy'' as the active stack goal. Verification
supplies the episode's single \textsc{Push}, the Venmo login prerequisite; after
$K$ consecutive search-parameter errors on the resulting lookup subgoal,
\textsc{Diagnose} attributes the failure to a wrong decomposition and
prescribes the one transition that follows from that attribution---a
\textsc{BacktrackTo} the parent, which in this episode restores the request
goal---instead of only revising the next free-form thought
(Figure~\ref{fig:case-diagnose}). After returning to that ancestor, the agent
reads the payment-request schema and executes
\texttt{create\_payment\_request} with the required amount, note, and privacy.
We do not claim that \textsc{Diagnose} itself detected the operator inversion
seen in ReflAct; the attributed failure was the lookup-parameter errors. The
contrast is still informative: ReflAct's free-form reflection can endorse a
fluent wrong operator, whereas VBL routes the mid-horizon failure to a typed
stack repair and then resumes from the outstanding request goal.

\begin{figure}[htbp]
\centering
\footnotesize
\setlength{\tabcolsep}{2pt}
\begin{tabular}{@{}>{\raggedright\arraybackslash}p{0.46\columnwidth}|
  >{\raggedright\arraybackslash}p{0.46\columnwidth}@{}}
\toprule
\multicolumn{2}{@{}p{0.94\columnwidth}@{}}{%
  \textbf{Task:} request \$13 from Stacy on Venmo
  (note: ``For yesterday's meal'', privacy: public)} \\
\midrule
\multicolumn{1}{c|}{\textbf{ReflAct (fail)}} &
\multicolumn{1}{c}{\textbf{VBL (succeed)}} \\
\midrule
login $\rightarrow$ find Stacy &
\textsc{Verify}$\rightarrow$\textsc{Push}(\texttt{login})
  $\rightarrow$ token \\[0.35em]
\textbf{Action:}
  \texttt{create\_transaction}
  (send money) &
\texttt{search\_friends} $\rightarrow$ arg error \\[0.15em]
 &
\fbox{\parbox{0.40\columnwidth}{\centering
  \textsc{Diagnose}$\rightarrow$\\
  \textsc{BacktrackTo}(parent)}} \\[0.15em]
Obs: ``Sent money.''
  mutates \texttt{Transaction} &
\textbf{Action:}
  \texttt{create\allowbreak\_payment\allowbreak\_request}
  (\$13, public note) \\[0.35em]
\textbf{Result:} $1/7$ tests &
\textbf{Result:} $7/7$ tests \\
\bottomrule
\end{tabular}
\caption{Additional case (task \texttt{024c982\_1}). ReflAct inverts the goal
with \texttt{create\_transaction}; VBL repairs the failed lookup on the stack
and keeps the request operator.}
\label{fig:case-diagnose}
\end{figure}

\section{Per-Type Gain Analysis}
\label{app:per-type}

Beyond aggregate success and control-operation usage, we examine \emph{where}
VBL's gains concentrate at the task-type level.

\paragraph{Gains concentrate on precondition-heavy task types.}
Table~\ref{tab:per-type} breaks ALFWorld down by task type. AppWorld exposes no
comparable type label, and ScienceWorld's $24$ types average fewer than nine
test episodes each---too few for a stable per-type estimate---so we report
ALFWorld quantitatively and summarize the ScienceWorld trend qualitatively.

VBL's gains fall mainly on the compositional types that must establish an
enabling state before the terminal placement:
\texttt{pick\_heat\_then\_place} ($+13.0$ points for Qwen3.5-9B, $+8.7$ for
DeepSeek-V3.2) and \texttt{pick\_clean\_then\_place} ($+6.5$ / $+12.9$), with the
single largest gain on \texttt{pick\_two\_obj\_and\_place} for DeepSeek-V3.2
($+35.3$). The already-easy \texttt{pick\_and\_place\_simple} type is saturated
for both methods. Two qualifications are worth stating explicitly. First, the
pattern is not uniform across backbones: for Qwen3.5-9B
\texttt{pick\_two\_obj\_and\_place} regresses by $5.8$ points and
\texttt{pick\_cool\_then\_place} is flat, even though both require an enabling
step, so the smaller backbone does not convert the extra control into gains
evenly. Second, Qwen3.5-9B's largest single gain is on
\texttt{look\_at\_obj\_in\_light} ($77.8\!\rightarrow\!100.0$), a type whose
difficulty lies in locating a light source rather than in a compositional state
change; this is still consistent with VBL supplying a missing prerequisite, but
the prerequisite is one of object discovery. On ScienceWorld the largest gains
likewise fall on procedure-heavy types such as \emph{boil}, \emph{melt},
\emph{chemistry-mix}, and \emph{find-living-thing}, whereas open-ended types with
no reliable procedure (e.g., \emph{grow-fruit}) remain hard for all methods.
Overall the gains are concentrated rather than uniform, which is consistent with
VBL helping by supplying missing preconditions, although the ALFWorld type labels
alone do not isolate that mechanism.

\begin{table}[htbp]
\centering
\small
\caption{ALFWorld success rate (\%) by task type. Gains often fall on
compositional types that require establishing an enabling state (e.g., clean or
heat) before the terminal action, though the pattern is not uniform across
backbones.}
\label{tab:per-type}
\setlength{\tabcolsep}{4pt}
\renewcommand{\arraystretch}{1.0}
\begin{tabular}{l cc cc}
\toprule
& \multicolumn{2}{c}{Qwen3.5-9B} & \multicolumn{2}{c}{DeepSeek-V3.2} \\
\cmidrule(lr){2-3}\cmidrule(lr){4-5}
ALFWorld task type & ReAct & VBL & ReAct & VBL \\
\midrule
look\_at\_obj\_in\_light        & 77.8 & 100.0 & 100.0 & 100.0 \\
pick\_and\_place\_simple        & 100.0 & 100.0 & 95.8 & 100.0 \\
pick\_clean\_then\_place        & 77.4 & 83.9 & 87.1 & 100.0 \\
pick\_cool\_then\_place         & 95.2 & 95.2 & 90.5 & 95.2 \\
pick\_heat\_then\_place         & 52.2 & 65.2 & 69.6 & 78.3 \\
pick\_two\_obj\_and\_place      & 88.2 & 82.4 & 58.8 & 94.1 \\
\bottomrule
\end{tabular}
\end{table}

\end{document}